 \documentclass[sigconf]{acmart}

\usepackage{amsmath,amssymb,amsfonts}
\usepackage{algorithmic}
\usepackage{graphicx}
\usepackage{textcomp}
\usepackage{xcolor}

\usepackage{booktabs}
\usepackage{algorithm}
\usepackage{algorithmic}

\AtBeginDocument{%
  \providecommand\BibTeX{{%
    \normalfont B\kern-0.5em{\scshape i\kern-0.25em b}\kern-0.8em\TeX}}}

\setcopyright{acmcopyright}
\copyrightyear{2018}
\acmYear{2018}
\acmDOI{XXXXXXX.XXXXXXX}

\acmConference[Conference acronym 'XX]{Make sure to enter the correct
  conference title from your rights confirmation emai}{June 03--05,
  2018}{Woodstock, NY}
\acmPrice{15.00}
\acmISBN{978-1-4503-XXXX-X/18/06}

\begin{document}

\title{Coarse-to-fine Knowledge Graph Domain Adaptation based on Distantly-supervised Iterative Training}

\author{Hongmin Cai}
\email{hmcai@scut.edu.cn}
\affiliation{%
  \institution{South China University of Technology}
  \country{China}
}

\author{Wenxiong Liao}
\email{cswxliao@mail.scut.edu.cn}
\affiliation{%
  \institution{South China University of Technology}
  \country{China}
}

\author{Zhengliang Liu}
\email{zl18864@uga.edu}
\affiliation{%
  \institution{University of Georgia}
  \country{USA}
}

\author{Yiyang Zhang}
\email{zyyinyourarea@163.com}
\affiliation{%
  \institution{South China University of Technology}
  \country{China}
}

\author{Xiaoke Huang}
\email{csxkhuang@mail.scut.edu.cn}
\affiliation{%
  \institution{South China University of Technology}
  \country{China}
}

\author{Siqi Ding}
\email{d15995291636@163.com}
\affiliation{%
  \institution{South China University of Technology}
  \country{China}
}

\author{Hui Ren}
\email{hren2@mgh.harvard.edu}
\affiliation{%
  \institution{Massachusetts General Hospital and Harvard Medical School}
  \country{USA}
}

\author{Zihao Wu}
\email{zihao.wu1@uga.edu}
\affiliation{%
  \institution{University of Georgia}
  \country{USA}
}

\author{Haixing Dai}
\email{hd54134@uga.edu}
\affiliation{%
  \institution{University of Georgia}
  \country{USA}
}

\author{Sheng Li}
\email{vga8uf@virginia.edu}
\affiliation{%
  \institution{University of Virginia}
  \country{USA}
}

\author{Lingfei Wu}
\email{Teddy.lfwu@gmail.com}
\affiliation{%
  \institution{Pinterest}
  \country{USA}
}

\author{Ninghao Liu}
\email{ninghao.liu@uga.edu}
\affiliation{%
  \institution{University of Georgia}
  \country{USA}
}

\author{Quanzheng Li}
\email{Li.Quanzheng@mgh.harvard.edu}
\affiliation{%
  \institution{Massachusetts General Hospital and Harvard Medical School}
  \country{USA}
}

\author{Tianming Liu}
\email{tliu@cs.uga.edu}
\affiliation{%
  \institution{University of Georgia}
  \country{USA}
}

\author{Xiang Li}
\email{XLI60@mgh.harvard.edu}
\affiliation{%
  \institution{Massachusetts General Hospital and Harvard Medical School}
  \country{USA}
}

\renewcommand{\shortauthors}{Trovato and Tobin, et al.}

\begin{abstract}
The knowledge graph (KG) is a highly needed basis to support the high-fidelity, high-interpretability modeling of various tasks in healthcare artificial intelligence. In this work, we focus on constructing an oncology knowledge graph that will be used in downstream cancer research and solution development. Modern supervised learning for knowledge graph construction requires a large amount of manually labeled data, which makes the process time-consuming and labor-intensive. Although there exists multiple research on named entity recognition and relation extraction based on distantly supervised learning, constructing a domain-specific knowledge graph from large collections of textual data without manual annotations is still an urgent problem to be solved. In response, we propose an integrated framework for adapting and re-learning knowledge graphs from a general domain (biomedical in our case) to a fine-defined domain (oncology). In this framework, we apply distant-supervision on cross-domain knowledge graph adaptation. Consequently, no manual data annotation is required to train the model. We introduce a novel iterative training strategy to facilitate the discovery of domain-specific named entities and triplets. Experimental results indicate that the proposed framework can perform domain adaptation and construction of knowledge graphs efficiently.
\end{abstract}





\maketitle

\section{Introduction}
In healthcare, the development of robust and interpretable clinical decision support systems and the corresponding research requires both a substantial amount of data and effective modeling of the medical domain knowledge \cite{ZHANG2020102324}. Knowledge graphs (KG) have been explored in healthcare to represent the underlying relationships representing the domain knowledge, including Unified Medical Language System (UMLS) \cite{lindberg1993unified} and Google Healthcare Knowledge Graph \cite{rotmensch2017learning}. In this study, we focus on developing the KG for oncology, an important branch of medicine that studies cancer treatment and prevention. Delineating the relationships between cancer sub-types, symptoms, comorbidities, genetic factors, and treatment with the oncology KG would provide a powerful basis for the downstream task in clinical decision support, such as patient diagnosis, prognosis, phenotyping, and treatment optimization.

However, existing approaches are not effective for constructing domain-specific KGs, especially for oncology, where limited access to oncological expertise hinders the supply of labeled training data. Insufficient labeled data typically leads to suboptimal performance. In fact, the dependence on sizeable training data significantly diminishes the real-world potential of data-driven KG construction methods based on supervised learning. In addition, although rule-based methods (based on resources such as Stanford CORE NLP) do not have stringent data demands, they typically suffer from suboptimal hand-crafted feature designs and the absence of helpful fine-grained connections to the domain data. Consequently, automatically constructing knowledge graphs directly from natural texts has attracted close attention from researchers in recent years \cite{kertkeidkachorn2017t2kg,rossanez2020kgen,stewart2020seq2kg}.

In order to address these challenges, we investigate the coarse-to-fine learning for constructing an oncology knowledge graph that leverages knowledge from general biomedical KGs, especially the distantly-supervised interactive training to achieve knowledge graph domain adaptation. In distantly-supervised learning, fine-domain KGs are derived from the general-domain KGs. For example, a biomedical KG that covers broad concepts and common sense knowledge in the biomedical domain can serve as the base KG for a specialized oncology KG. Therefore, the KG in the coarse domain can be used as a knowledge base for distant supervision, thus avoiding the need for extensive manual annotations. However, only using the KG of the coarse domain as the knowledge base might limit the model's ability to discover domain-specific named entities and triples in the fine domain, further limiting the construction of the fine domain KG. Thus in this paper, we propose a novel coarse-to-fine knowledge graph domain adaptation (KGDA) framework. Our KGDA framework utilizes an iterative training strategy to enhance the model's ability to discover fine-domain entities and triples, thereby facilitating fast and effective coarse-to-fine KG domain adaptation. 

% In the construction of fine-domain KG  scenarios, there are usually some existing resources available, such as biomedical KGs in coarse domains, which generally cover broader concepts and more commonsense knowledge. When constructing the oncology KG, the biomedical KG is thus available. However, few studies have focused on adapting KG from the coarse domain (e.g., biomedical) to the fine domain (e.g., oncology) where a large collection of unlabeled textual data are available, which motivates the work in this paper.
Overall, the contributions of our work are as follows:
\begin{itemize}
	\item An integrated framework for adapting and re-learning KG from coarse-domain to fine-domain is proposed. As a case study, the biomedical domain and oncology domain are considered the coarse domain and fine domain, respectively.
	\item Our model does not require human annotated samples with distant-supervision for cross-domain KG adaptation, and  the iterative training strategy is applied to discovering domain-specific named entities and new triples.
	\item The proposed method can be adapted to various pre-trained language models (PLMs) and can be easily applied to different coarse-to-fine KGDA tasks. It is so far the simplest data-driven approach for learning a KG from free text data, with the help of the coarse domain KG. 
	\item Experimental results demonstrate the effectiveness of the proposed KGDA framework. We will release the source code and the data used in this work to fuel further research. The constructed oncology KG will be hosted as a web service to be used by the general public.
	
\end{itemize}

\section{Background and Related work}
Automatic KG construction from text generally involves two primitive steps: named entity recognition (NER) and relation extraction (RE). Named entity recognition aims to identify the types of entities mentioned in text sequences, such as people, places, etc. in the open domain; or diseases, medicine, disease symptoms, etc. in the biomedical domain. Relation extraction is also known as triplet extraction, which aims to identify the relationship between two entities, such as the birthplace relationship between people and places, or the therapeutic relationship between drugs and diseases in the biomedical domain. NER and RE are the necessary steps for information extraction to construct KG from text. 

Distant supervision \cite{smirnova2018relation} is an intuitive way to transfer general-domain KG to fine domains. Distant-supervision provides labels for data with the help of an external knowledge base, which saves the time of manual labeling. For distantly-supervised NER, we can build distant labels by matching unlabeled sentences with external semantic dictionaries or knowledge bases. The matching strategies usually include string matching \cite{zhao2019construction}, regular expressions \cite{fries2017swellshark}, and some heuristic rules. The distantly-supervised RE holds an assumption \cite{mintz2009distant}: if two entities participate in a relation, then any sentence that contains those two entities might express that relation. Following this assumption, any sentence mentioning a pair of entities that have a relation according to the knowledge base will be labeled with this relation \cite{smirnova2018relation}. 
\subsection{Pipeline-based methods for KG construction}
The pipeline-based methods apply carefully-crafted linguistic and statistical patterns to extract the co-occurred noun phrases as triples. There are many off-the-shelf  toolkits available, for example, Stanford CoreNLP \cite{manning2014stanford}, NLTK \cite{thanaki2017python}, and spaCy, which can be used for the NER tasks; Reveb \cite{fader2011identifying}, OLLIE \cite{schmitz2012open}, and Stanford OpenIE \cite{angeli2015leveraging} can be used for the information extraction task. There have been multiple pipelines \cite{mehta2019scalable,rossanez2020kgen} developed as well, consisting of modules targeting different functionalities needed for the KG construction.  However, the pre-defined rules of off-the-shelf  toolkits are generally tailored to specific domains, such methods are not domain-agnostic, and a new set of rules will be needed for a new domain. 
\subsection{Data-driven methods for KG construction}
With the development of representation learning in language models, researchers began to apply data-driven models to solve the KG construction tasks. Based on how the model is trained, such work can be divided into three categories: fully-supervised methods \cite{zhao2019construction, li2022medukg}, semi-supervised methods \cite{zahera2021asset}, and weakly-supervised methods \cite{yu2021domain}. We will introduce the methods of fully-supervised and weakly-supervised in this section. Specifically, the NER, RE, and entity linking tasks in the KG construction pipeline can all be solved by fully-supervised learning methods such as long short-term memory neural network (LSTM) \cite{hochreiter1997long,zeng2017lstm}. Graph neural network methods have also been applied for domain-specific NER tasks \cite{chen2021explicitly} and document-level RE \cite{zhang2020document}. The bidirectional encoder representation from transformers (BERT) \cite{kenton2019bert}, a widely-used pretrained language model (PLM), can also tackle the NER \cite{jia2020entity}, RE \cite{roy2021incorporating}, and entity linking \cite{li2022lp} tasks. While the advancement of deep learning-based methods has greatly improved the effectiveness of KG construction, fully-supervised learning requires a large amount of human-annotated data text. Furthermore, the annotation can only be domain-specific, making it difficult to transfer the KG construction work to a new domain, and ultimately limiting the scalability and efficiency of the research in KG.

On the other hand, distant supervision, a weakly supervised learning method, can replace manual annotation with an existing and remote knowledge base. Previous studies have applied remote supervised learning to deal with NER \cite{zheng2021distantly}, and RE \cite{wei2021distantly,zhang2021readsre} tasks. Thus in this work, we adopted the distant-supervision scheme in the proposed KGDA framework. It should be noted that KG of the coarse domain (e.g., biomedical) generally will not contain the complete knowledge of its finer sub-domains (e.g., oncology). So when we use the coarse-domain KG for distant supervision, labels of the target domain will be limited by the source domain, making it less effective to discover new knowledge. To address this issue, we introduced an iterative strategy to gradually update the model via distant supervision while at the same time using the partially-trained model to discover new entities and relations from the data of the target fine domain. 

\begin{figure*}[t]
	\centering
	\includegraphics[width=0.95\textwidth]{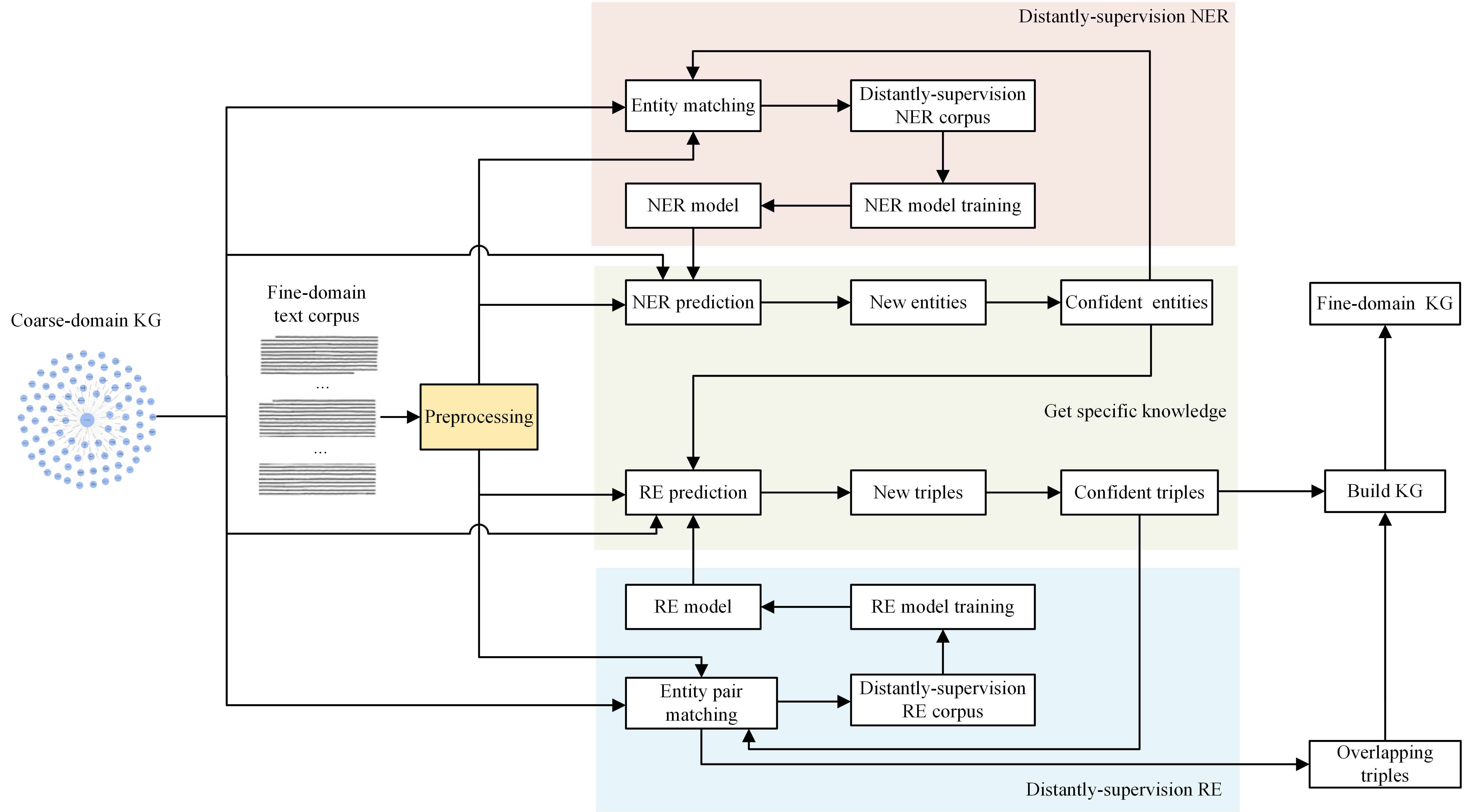} % Reduce the figure size so that it is slightly narrower than the column.
	\caption{The overall framework of iterative training KGDA.}
	\label{fig1}
\end{figure*}

\section{Methodology}
\subsection{Notation and task definition}
An unstructured sentence $s = [w_1,w_2,w_3,...,w_n]$ indicates a sequence of tokens, where $n$ is its length. A dataset $\mathbb{D}$ is a collection of unstructured sentences (i.e. $\mathbb{D} = \{s_1,s_2,s_3,..,s_m\}$). The knowledge graph, denoted as $\mathbb{K}$, is a collection of triples $t = (e_i,r_j,e_k)$, where $e_i \in \mathbb{E}$ and $e_k \in \mathbb{E}$ are the head entity and the tail entity respectively, and $r_j \in \mathbb{V}$ is the relation between $e_i$ and $e_k$. Here we denote coarse-domain KG as $\mathbb{K}_c$ and fine-domain KG as $\mathbb{K}_f$.

In a typical scenario of KG domain adaptation, we will have an existing coarse-domain KG and a large amount of unlabeled text in the fine domain. For example, when constructing the oncology KG, we can utilize the existing biomedical KG and collect oncology-related literature as unlabeled text. KG constructed from the fine domain data would then include overlapping triples with the coarse-domain KG and new triples representing domain-specific knowledge. Specifically, the fine-domain KG contains the following three types of triples:

\begin{itemize}
	\item \textbf{Overlapping triples $\mathbb{T}_O$}: Triples that also existed in the coarse-domain KG, indicating knowledge overlapping between the coarse and fine domains. 
	
	\item \textbf{Triples of new relations but overlapping entities $\mathbb{T}_R$}: Triples with both entity pairs existing in the coarse-domain KG but no indicated relationships between these entity pairs.
	
	\item \textbf{Triples of new entities $\mathbb{T}_E$}: Triples with at least one entity not existing in the coarse-domain KG. Consequently, the relationship is also unknown in the coarse domain.
\end{itemize}

Both $\mathbb{T}_R$ and $\mathbb{T}_E$ belong to the specific knowledge of the fine domain. The goal of the coarse-to-fine KGDA task is to adapt the KG from the coarse domain to the fine domain and leverage the knowledge from the coarse domain to guide the mining of new knowledge specific to the fine domain. Finally, we will keep the definition of entity types and relation types from coarse-domain KG when constructing the fine-domain KG.

\subsection{Iterative training framework}
While it is trivial to identify the overlapping entities $\mathbb{E}_O$ and triples $\mathbb{T}_O$ by distant supervision, if the NER and RE models are trained on the entire corpus, they will not be able to recognize the fine domain-specific named entities and triples ($\mathbb{T}_R$ and $\mathbb{T}_E$). Because the distant-supervision labels are generated by matching $\mathbb{K}_c$. Thus we introduce an iterative training strategy to construct $\mathbb{T}_R$ and $\mathbb{T}_E$ from the text and adapt the knowledge from $\mathbb{K}_c$ to $\mathbb{K}_f$.

The overall framework of the iterative training scheme is shown in Fig. \ref{fig1}, and the detailed pseudo code can be found in Algorithm \ref{alg:algorithm1}. Rather than performing distant-supervision training on the whole unlabeled text corpus, the core mechanism of the proposed iterative training is to split the whole unlabeled dataset into $n$ sub-datasets without intersection. Before building distant-supervision corpus, the trained model is used to predict the text corpus for getting specific knowledge of fine-domain, which is conducive to mining $\mathbb{T}_R$ and $\mathbb{T}_E$ of the fine-domain.

As shown in Figure \ref{fig1}, firstly, it is necessary to preprocess the acquired text corpus in the fine domain. Preprocessing operations include: handling special characters, word segmentation, filtering sentences using human-defined rules (such as sentence length), etc. Then, our framework involves two neural network models: NER model and RE model. We replace the PLM’s output layer with a classifier head as  NER model $model_N$ and fine-tune it by minimizing the cross-entropy loss on distant-supervision NER corpus. Additionally, we apply the BIO scheme \cite{li2012joint} to generate NER sequence labels. For the RE task, we use the template to generate distant-supervision samples. The template we adopted is "[CLS] $head\ entity$ ($head\ entity\ type$) [SEP] $tail\ entity$ ($tail\ entity\ type$) [SEP] $sentence$". The RE model $mode_R$ is defined as a PLM with a fully connected layer as a relation classifier. The feature of special token [CLS]  fed into this fully connected layer and fine-tune $mode_R$ by minimizing the cross-entropy loss on distant-supervision RE corpus.

We summarize the steps to achieve KGDA in Algorithm  \ref{alg:algorithm1}. For the first parts of the text corpus $\mathbb{D}_1$, the distant-supervision method is applied to construct the NER training corpus $corp_N$ and RE training corpus $corp_R$ , and the NER model $model_N$ and RE model $mode_R$ are trained based on corpus $corp_N$ and $corp_R$, respectively. For other part of the text corpus $\mathbb{D}_i$,  we apply the previously trained $model_N$ and $mode_R$ to extract the entities and triples in the fine-domain, and select the  high confidence entities $\mathbb{E}_{conf}$ and  high confidence triples $\mathbb{T}_{conf}$ as the specific knowledge of the fine-domain (line 7). Then, we take $\mathbb{K}_c$, $\mathbb{E}_{conf}$, and $\mathbb{T}_{conf}$  as the external knowledge base for constructing distant-supervision $corp_N$ and $corp_R$ (line 8). Finally, we use overlapping triples $\mathbb{T}_O$ and high-confidence triples $\mathbb{T}_{conf}$  to construct a knowledge graph of fine domains (line 17).

Next, we show the details of \textit{get\_distant\_corpus} in Algorithm \ref{alg:algorithm2} and \textit{get\_specific\_ knowledge} in Algorithm \ref{alg:algorithm3}.

\begin{algorithm}[tb]
	\caption{Iterative training KGDA framework}
	\label{alg:algorithm1}
	\textbf{Input}: Text corpus $\mathbb{D} = \{\mathbb{D}_1,\mathbb{D}_2,...,\mathbb{D}_n\}$, coarse-domain KG $\mathbb{K}_c$, out-of-domain words $\mathbb{W}_O$ \\
	\textbf{Parameter}: Initialized NER model $model_N$, initialized RE model $model_R$ \\
	\textbf{Output}:  fine-domain kg $\mathbb{K}_f$
	\begin{algorithmic}[1] %[1] enables line numbers
		\STATE Let new entities  $\mathbb{E}_{new}=\{\}$ , new entities  with high confidence $\mathbb{E}_{conf}=\{\}$ , new triples  $\mathbb{T}_{new}=\{\}$ , new triples  with high confidence $\mathbb{T}_{conf}=\{\}$ .
		\STATE $corp_N$, $corp_R$, $\mathbb{E}_O$, $\mathbb{T}_O$= build\_distant\_corpus( $\mathbb{D}_1$, $\mathbb{K}_c$, $\mathbb{E}_{conf}$, $\mathbb{T}_{conf}$, $\mathbb{W}_O$ )
		\STATE train\_NER($model_N$,  $corp_N$)
		\STATE train\_RE($model_R$,  $corp_R$)
		\STATE $i = 2 $
		\WHILE{$i <= n$}
		\STATE $\mathbb{E}_{new}$, $\mathbb{E}_{conf}$, $\mathbb{T}_{new}$, $\mathbb{T}_{conf}$ = get\_specific\_knowledge($\mathbb{D}_i$, $\mathbb{K}_c$, $\mathbb{E}_{new}$, $\mathbb{E}_{conf}$, $\mathbb{T}_{new}$, $\mathbb{T}_{conf}$ )
		\STATE $corp_N^{'}$, $corp_R^{'}$, $\mathbb{E}_O^{'}$, $\mathbb{T}_O^{'}$= get\_distant\_corpus( $\mathbb{D}_i$, $\mathbb{K}_c$, $\mathbb{E}_{conf}$, $\mathbb{T}_{conf}$, $\mathbb{W}_O$)
		\STATE $corp_N$ = $corp_N \cup corp_N^{'}$
		\STATE $corp_R$ = $corp_R \cup corp_R^{'}$
		\STATE$\mathbb{E}_O$ = $\mathbb{E}_O \cup \mathbb{E}_O^{'}$
		\STATE$\mathbb{T}_O$ = $\mathbb{T}_O \cup \mathbb{T}_O^{'}$
		\STATE train\_NER($model_N$,  $corp_N$)
		\STATE train\_RE($model_R$,  $corp_R$)
		
		\STATE $i$ = $i + 1 $
		\ENDWHILE
		\STATE $\mathbb{K}_f$ = build\_kg($\mathbb{T}_{O}$ , $\mathbb{T}_{conf}$)
		\STATE \textbf{return} $\mathbb{K}_f$
	\end{algorithmic}
\end{algorithm}

\begin{algorithm}[tb]
	\caption{Constructing distantly-supervised corpus}
	\label{alg:algorithm2}
	\textbf{Input}: A part of text corpus text corpus $\mathbb{D}_i$, coarse-domain KG $\mathbb{K}_c$, new entities with high confidence $\mathbb{E}_{conf}$, new triples  with high confidence $\mathbb{T}_{conf}$, out-of-domain words $\mathbb{W}_O$ \\
	\textbf{Parameter}:  negative sample ratio ${ratio_n}$ ,  out-of-domain sample ratio ${ratio_o}$ \\
	\textbf{Output}: Distant-supervision NER corpus $corp_N$, distant-supervision RE corpus $corp_R$, overlapping entities $\mathbb{E}_O$, overlapping triples $\mathbb{T}_O$  
	\begin{algorithmic}[1] %[1] enables line numbers
		\STATE Let  $corp_E=\{\}$,  $corp_R=\{\}$, $\mathbb{E}_O=\{\}$, $\mathbb{T}_O=\{\}$.
		\STATE sentence\_num = len( $\mathbb{D}_i$)
		\STATE $j = 1$
		\WHILE{j$<=$sentence\_num}
		\STATE $entities$ = entity\_matching( $\mathbb{D}_i^{j}$, $\mathbb{K}_c$, $\mathbb{E}_{conf}$)
		\STATE $\mathbb{E}_O$ = $\mathbb{E}_O \cup entities$
		\STATE  $corp_N$ =  $corp_N \cup$ build\_NER\_sample( $\mathbb{D}_i^{j}$, $entities$)
		
		\STATE $triples_k$,$triples_c$ = entity\_pair\_matching( $\mathbb{D}_i^{j}$, $\mathbb{K}_c$, $\mathbb{T}_{conf}$)
		\STATE  $triples$ = $triples_k \cup triples_c$
		\STATE $triples_n$ = get\_negative\_triples( $\mathbb{D}_i^{j}$, $\mathbb{W}_O$, $triples$, ${ratio_n}$, ${ratio_o}$)
		\STATE  $corp_R$ = $corp_R  \cup $ get\_samples($triples$)
		\STATE  $corp_R$ = $corp_R  \cup $ get\_samples($triples_n$)
		\STATE $\mathbb{T}_O$ = $\mathbb{T}_O \cup triples_k $  
		\STATE $j = j + 1$
		\ENDWHILE
		\STATE \textbf{return} $corp_N$, $corp_R$, $\mathbb{E}_O$, $\mathbb{T}_O$ 
	\end{algorithmic}
\end{algorithm}

\begin{algorithm}[tb]
	\caption{Discovering fine-domain specific knowledge}
	\label{alg:algorithm3}
	\textbf{Input}: A part of text corpus text corpus $\mathbb{D}_i$, coarse-domain KG $\mathbb{K}_c$,   new entities  $\mathbb{E}_{new}$ , new entities  with high confidence $\mathbb{E}_{conf}$ , new triples  $\mathbb{T}_{new}$ , new triples  with high confidence $\mathbb{T}_{conf}$ \\
	\textbf{Parameter}: NER model $model_N$, RE model $model_R$, probability threshold of the entity $th_{pe}$, frequency threshold of the entity $th_{fe}$, probability threshold of the triple $th_{pt}$, frequency threshold of the triple $th_{ft}$ \\
	\textbf{Output}: $\mathbb{E}_{new}$, $\mathbb{E}_{conf}$, $\mathbb{T}_{new}$, $\mathbb{T}_{conf}$
	\begin{algorithmic}[1] %[1] enables line numbers
		\STATE Let $corp_E=\{\}$, $corp_R=\{\}$, $\mathbb{E}_O=\{\}$, $\mathbb{T}_O=\{\}$.
		\STATE sentence\_num = len( $\mathbb{D}_i$)
		\STATE $j = 1$
		\WHILE{$j<=sentence\_num$}
		\STATE $entities$ = NER\_prediction( $\mathbb{D}_i^{j}$, $model_N$)
		\STATE $entities$ = get\_new\_entities( $entities$, $\mathbb{K}_c$)
		\STATE $\mathbb{E}_{new}$ = merge\_entity( $\mathbb{E}_{new}$, $entities$)
		\STATE $j = j + 1$
		\ENDWHILE
		\STATE $\mathbb{E}_{conf}$ = get\_confidence\_entity( $\mathbb{E}_{new}$, $th_{pe}$, $th_{fe}$)
		
		\STATE $j = 1$
		\WHILE{$j<=sentence\_num$}
		\STATE $entities$ = entity\_matching( $\mathbb{D}_i^{j}$, $\mathbb{K}_c$, $\mathbb{E}_{conf}$ )
		\STATE $pairs$ = enumerate\_pairs( $entities$ )
		\STATE $pairs$ = get\_new\_pairs( $pairs$, $\mathbb{K}_c$ )
		\STATE $triples$ = RE\_prediction( $\mathbb{D}_i^{j}$, $pairs$ , $model_R$)
		\STATE $\mathbb{T}_{new}$ = merge\_triple( $\mathbb{T}_{new}$, $triples$ )
		\STATE $j = j + 1$
		\ENDWHILE
		\STATE $\mathbb{T}_{conf}$ = get\_confidence\_triple( $\mathbb{T}_{new}$, $th_{pt}$, $th_{pt}$ )
		
		\STATE \textbf{return} $\mathbb{E}_{new}$, $\mathbb{E}_{conf}$, $\mathbb{T}_{new}$, $\mathbb{T}_{conf}$
	\end{algorithmic}
\end{algorithm}

\subsection{Constructing distantly-supervised corpus}
Through distant-supervision, we can only match entity pairs that have a relationship and use them as positive samples. We then construct negative samples with NULL relationship by the following two schemes: 1) randomly sampling two entities which have no relationship as defined in the coarse-domain; 2) randomly sampling a word from out-of-domain words (i.e., a word that is not an entity as defined in the coarse domain) $\mathbb{W}_O$ as one of the entities. The parameter ${ratio_n}$ controls the ratio of negative samples (constructed by either schemes) to the total sample size. The parameter ${ratio_o}$ controls the ratio of entity pairs constructed by the second scheme (i.e., via sampling the words outside the domain) to the size of negative samples, respectively. 

In addition to the $\mathbb{K}_c$ in the source domain, we use both $\mathbb{K}_c$, $\mathbb{E}_{conf}$, and $\mathbb{T}_{conf}$ as knowledge bases for constructing the remotely supervised corpus. This would ensure that the NER and RE models can identify the overlapping knowledge between $\mathbb{K}_c$ and $\mathbb{K}_f$, while at the same time be guided to discover the new knowledge specific to the fine domain. 

As shown in Algorithm \ref{alg:algorithm2}, for building the distantly-supervised NER corpus $corp_N$, the sentence $\mathbb{D}_i^{j}$ is firstly string-matched with the knowledge bases $\mathbb{K}_c$ and $\mathbb{E}_{conf}$ to extract the entities in the sentence (line 5). Afterward, the matched entities are merged into overlapping entities $\mathbb{E}_O$, and the NER label sequences are generated through the BIO strategy to merge into $corp_N$ (line 6 and 7). For building the distantly-supervised RE corpus $corp_R$, we firstly take $\mathbb{K}_c$ and $\mathbb{T}_{conf}$ as knowledge bases and use entity pair matching to match the triples $triples_k$ based on $\mathbb{K}_c$ and the triples $triples_c$ based on $\mathbb{T}_{conf}$ appearing in the sentence $\mathbb{D}_i^{j}$ (line 8). We then build negative triples with parameters ${ratio_n}$ and ${ratio_o}$ (line 10). Finally, we construct the RE corpus based on the triples $triples$, $triples_n$ and corresponding sentences through a pre-defined relationship sample template (line 11 and 12).

\subsection{Discovering fine-domain specific knowledge}

Recall that in the proposed iterative training framework, the whole unlabeled dataset is divided into $n$ sub-dataset $\mathbb{D}_i,i=1...n$, the fine-domain specific knowledge discovery will be performed on each sub-dataset except the first one $\mathbb{D}_i,i=2...n$ (line 5 to 16 in Algorithm \ref{alg:algorithm1}). For each new sub-dataset $\mathbb{D}_i,i=2...n$, we will use the previously-updated models $model_N$ and $model_R$ to predict the new entities and triples. Afterward, the sub-dataset will be used for updating $model_N$ and $model_R$ via distantly-supervised training. As noisy or incorrect entities and triples could be discovered during this procedure, we developed a filtering mechanism only to keep the entities and triples with higher confidence. Specifically, we design the rules for filtering the discovered entities and triples by: 1) probability of the new entities and triples predicted by the corresponding models should be greater than pre-defined thresholds $th_{pe}$ and $th_{pt}$, respectively; 2) cumulative frequency of the new entities and triples discovered from datasets $\mathbb{D}_2$ to $\mathbb{D}_i$ should be greater than the pre-defined thresholds $th_{fe}$ and $th_{ft}$, respectively.

As shown in Algorithm \ref{alg:algorithm3}, for discovering new entities $\mathbb{E}_{new}$, we will apply the trained $model_N$ on dataset $\mathbb{D}_i$ and obtain $entities$ that are disjoint with $\mathbb{K}_c$ (line 5 and 6). Then, we will merge $entities$ with the previously-discovered entity set $\mathbb{E}_{new}$ (line 7). Finally, we will select the "high-confident" entity as $\mathbb{E}_{conf}$ based on the mechanism above by the prediction probability and cumulative frequency (line 10). For the discovery of new triples $\mathbb{T}_{new}$, we will enumerate entity pairs that are disjoint with the $\mathbb{K}_c$ (line 13 - 15). We will then use the trained RE model and the predefined sample template to predict the relationship of the entity pairs and delete the triples whose predicted relationship is NULL (line 16). Other processing is similar to the discovery of new entities.

After Algorithm \ref{alg:algorithm3}, discovered entities specific to the fine domain are stored in $\mathbb{E}_{conf}$. Discovered triples $\mathbb{T}_R$ (new relation, overlapping entity) and $\mathbb{T}_E$ (new relation, new entity) are stored in $\mathbb{T}_{conf}$. In the next iteration, Algorithm \ref{alg:algorithm2} will then use the updated $\mathbb{E}_{conf}$ and $\mathbb{T}_{conf}$ for building distant-supervision corpus. Such iterative design can facilitate the interoperability between the two competing tasks based on a fixed number of unannotated data samples in the fine target domain: distantly-supervised training of the NER and RE models versus the discovery of new knowledge using the trained NER and RE models, thus improve the efficiency of performing KG domain adaptation and construction without any annotation.

\section{Experiments}
In this work, we used the adaptation of KG from the biomedical domain (coarse) to the oncology domain (fine) as an example to demonstrate the workflow of the KGDA framework, as well as to evaluate its effectiveness in practice. Implementation details of the experiment are also provided, along with the publicly-available data and the containerized environment in the released source code, for easy replication of the experiment and the development of other KG methods. 

\subsection{Dataset}

The dataset we used in this paper is released by \cite{rezayi2022clinicalradiobert}. The source data is 
 downloaded  from 12 international journals in the oncology domain. PDF files of the papers were cleaned and converted to sentences. In total, we select 240,000 paragraphs as the unlabeled text corpus of the oncology domain $\mathbb{D}$. The coarse-domain KG $\mathbb{K}_c$ used in this work is the biomedical KG\footnote{https://idea.edu.cn/bios.html}, defines 18 entity types (anatomy, neoplastic process, microorganism, eukaryote, physiology, chemical or drug, diagnostic procedure, laboratory procedure, research activity or technique, therapeutic or preventive procedure, medical device, research device, pathology, disease or syndrome, anatomical abnormality, mental or behavioral dysfunction, injury or poisoning and sign, symptom or finding) and 19 relationship types (is\_a, reverse\_is\_a, is\_part\_of, reverse\_is\_part\_of, may\_treat, reverse\_may\_treat, found\_in, reverse\_found\_in, may\_cause, reverse\_may\_cause, expressed\_in, is\_expression\_of, encodes, encoded\_by, significant\_drug\_interaction, involved\_in\_biological\_process, biological\_process\_involves, is\_active\_ingredient\_in, has\_active\_ingredient), including ~5.2 million English entities and ~7.34 million triples. 

\subsection{Evaluation}

Similar to the previous works \cite{mintz2009distant}, we evaluate our method in two schemes: held-out evaluation and manual evaluation. For the held-out evaluation, we reserved a part of the text corpus of $\mathbb{D}$ as the test set. During the testing, we then compared the prediction results of the NER and RE models with the labels matched with $\mathbb{K}_c$, and calculated the precision, recall, and F1 of the held-out dataset. Specifically, we use seqvel\footnote{https://github.com/chakki-works/seqeval} to evaluate the micro average precision, recall, F1 of NER. When evaluating the RE model, we perform relation classification prediction on the triples existing in $\mathbb{K}_c$ and corresponding entity pairs appearing in the held-out corpus. Finally, weighted average precision, recall, and F1 from the held-out evaluation will be reported.

As the labels of testing samples in the held-out evaluation are all inferred by distant supervision from the coarse domain, such scheme can only evaluate whether the trained model can capture the knowledge in the coarse domain, but cannot evaluate the ability of the models to discover new knowledge in the fine-domain. Therefore, we also adopted the manual evaluation scheme, consisting of the evaluations of: 1) the entities specific to fine domain $\mathbb{E}_{conf}$, which are not presented in $\mathbb{K}_c$; 2) the triples of new relations $\mathbb{T}_R$; 3) the triples of new entities $\mathbb{T}_E$. We randomly sampled 50 cases of $\mathbb{E}_{conf}$, $\mathbb{T}_R$, and $\mathbb{T}_E$ respectively, then asked one physician to manually label them for whether the entities and triples are correct. As the number of name entities and triples instances that are expressed in the corpus is unknown, we cannot estimate the recall of fine-domain KG. Therefore, we only show the precision of $\mathbb{E}_{conf}$, $\mathbb{T}_R$, and $\mathbb{T}_E$. We fully recognize that the discovery of new knowledge in the fine-domain is an indispensable task for this work and we are recruiting more medical experts to conduct human reader study and performance evaluation for the proposed model. 

\subsection{Implementation settings}
We divide the corpus $\mathbb{D}$ into six equal subsets, and each subset contains around 40,000 sentences. We used $\mathbb{D}_1$ to $\mathbb{D}_5$ for model training and KG construction. We reserved $\mathbb{D}_6$ for held-out evaluation. We tested BERT \cite{kenton2019bert}, Bio\_ClinicalBERT \cite{alsentzer2019publicly}, biomed\_RoBERTa \cite{gururangan2020don} for initializing NER and RE models. Our experiments were run on an Ubuntu system computer with 4 NVIDIA A100 graphics cards. The learning rate, batch size, and epochs are set as 2E-05, 20, and 4, respectively. Hyperparameters $th_{fe}$,$th_{pe}$,$th_{ft}$,$th_{pt}$ are set as 2, 0.95, 3, 0.97. The parameters ${ratio_n}$ and ${ratio_o}$ that control negative sampling are set to 0.2 and 0.3.

\subsection{Held-out evaluation}

\begin{table}[h]
	\centering
	%\resizebox{.95\columnwidth}{!}{
	\begin{tabular}{llll}
        \hline
        models            & precision      & recall         & F1             \\ \hline
        BERT              & 0.908          & 0.900          & 0.904          \\
        Bio\_ClinicalBERT & \textbf{0.909} & 0.895          & 0.902          \\
        biomed\_RoBERTa   & 0.908          & \textbf{0.901} & \textbf{0.905} \\ \hline
        \end{tabular}
	\caption{Held-out evaluation of NER model.}
	\label{table1}
\end{table}

\begin{table}[h]
	\centering
	%\resizebox{.95\columnwidth}{!}{
	\begin{tabular}{llll}
        \hline
        \textbf{models}   & precision      & recall         & F1             \\ \hline
        BERT              & 0.987          & 0.949          & 0.967          \\
        Bio\_ClinicalBERT & \textbf{0.988} & 0.957          & \textbf{0.972} \\
        biomed\_RoBERTa   & 0.987          & \textbf{0.959} & \textbf{0.972} \\ \hline
        \end{tabular}
	\caption{Held-out evaluation of RE model.}
	\label{table2}
\end{table}

The results of the NER and RE models evaluated by the held-out dataset are shown in Table \ref{table1} and Table \ref{table2}, respectively. The KGDA frameworks initialized by the three pre-trained language models (BERT, Bio\_ClinicalBERT, and biomed\_RoBERTa) all show good performance in held-out evaluations, demonstrating the robustness of our framework. Because Bio\_ClinicalBERT and biomed\_RoBERTa are pre-trained in biomedical data sets, their performance is better than BERT.

\subsection{Manual evaluation}

\begin{figure*}[t]
	\centering
	\includegraphics[width=0.75\textwidth]{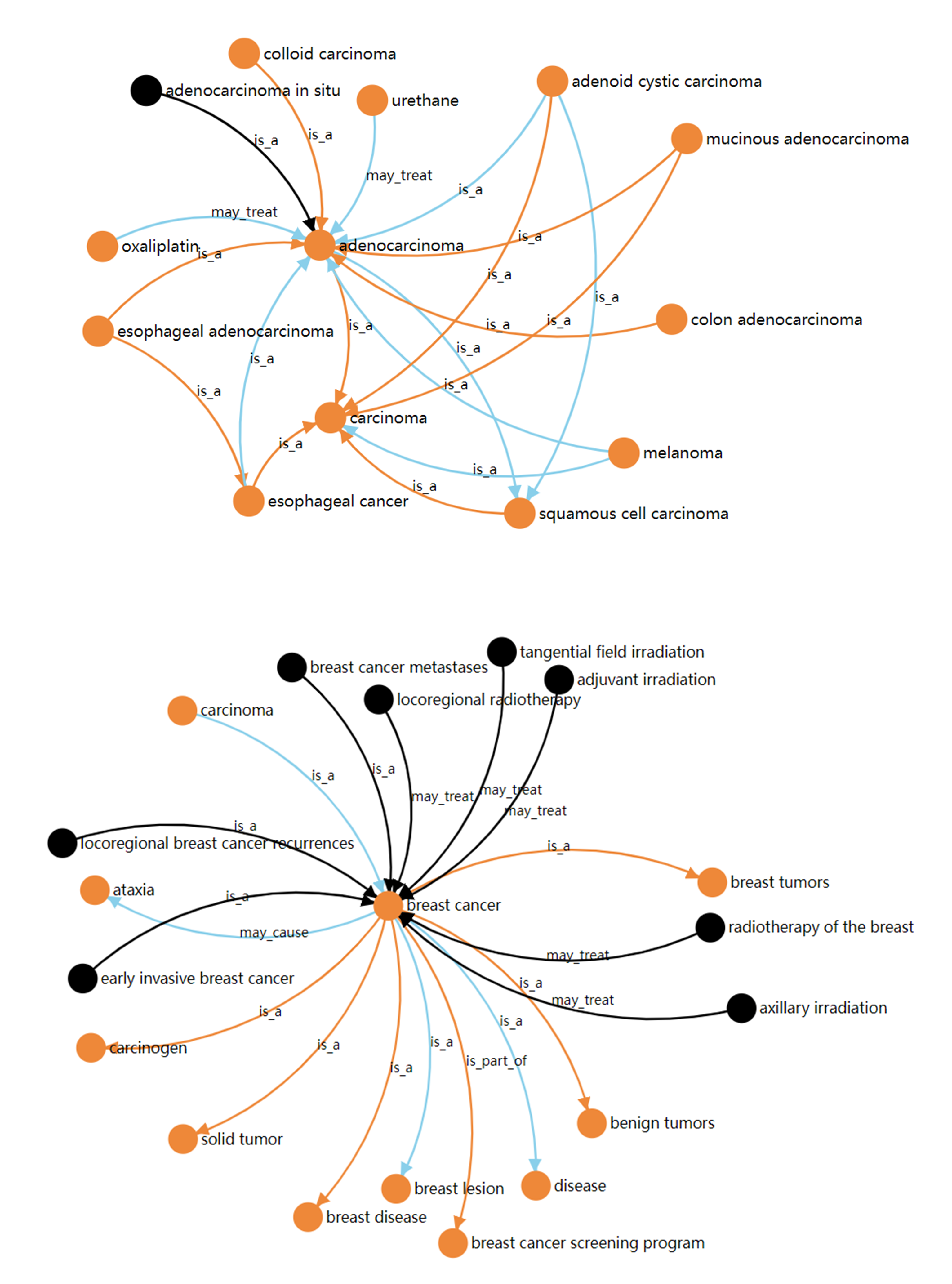} % Reduce the figure size so that it is slightly narrower than the column.
	\caption{knowledge graph of the \textit{adenocarcinoma} and \textit{breast cancer}. The orange edges represent the overlapping triples  $\mathbb{T}_O$, the blue edges denote the triples of new relations $\mathbb{T}_R$, and the black edges denote the triples of new entities $\mathbb{T}_E$, while the black node represents the new entity. It should be noted that in order to facilitate the display, we only show some associated triples, not all.}
	\label{fig2}
\end{figure*}

\begin{table}[h]
	\centering
	%\resizebox{1.0\columnwidth}{!}{
	\begin{tabular}{llllll}
		\hline
		models            & \#$\mathbb{E}_{O}$ & \#$\mathbb{T}_{O}$ & \#$\mathbb{E}_{conf}$ & \#$\mathbb{T}_R$ & \#$\mathbb{T}_E$ \\ \hline
		BERT              & 86741        & 26467       & 1378           & 36010         & 1178                   \\
            Bio\_ClinicalBERT & 86801        & 26353       & 1541           & 39588         & 1413                   \\
            biomed\_RoBERTa   & 86821        & 26448       & 1444           & 37637         & 1110                   \\ \hline
	\end{tabular}
	
	\caption{The number of entities and triples.}
	\label{table3}
\end{table}

\begin{table}[h]
	\centering
	%\resizebox{.95\columnwidth}{!}{
	\begin{tabular}{llll}
		\hline
		models            & $\mathbb{E}_{conf}$ & $\mathbb{T}_R$ & $\mathbb{T}_E$ \\ \hline
		BERT              & 0.90           & 0.58          & 0.70          \\
            Bio\_ClinicalBERT & 0.90           & 0.66          & 0.62          \\
            biomed\_RoBERTa   & \textbf{0.94}  & \textbf{0.76} & \textbf{0.74} \\ \hline
	\end{tabular}
	\caption{results of manual evaluations.}
	\label{table4}
\end{table}

The number of all discovered entities ($\mathbb{E}_{O}$), triples ($\mathbb{T}_{O}$), new entities with high confidence ($\mathbb{E}_{conf}$), triples representing new relations with overlapping entities ($\mathbb{T}_R$), and triples representing new relations with new entities ($\mathbb{T}_E$) are shown in Table \ref{table3}, with each row belonging to one pre-trained language models used. Numbers of $\mathbb{E}_{O}$ and $\mathbb{T}_{O}$ have minor differences among different pre-trained language models, possibly due to the conflicts in strings matching of knowledge bases. $\mathbb{E}_{conf}$, $\mathbb{T}_R$, and  $\mathbb{T}_E$ represent specific knowledge of the fine domain. We sampled 50 cases from $\mathbb{E}_{conf}$, $\mathbb{T}_R$, and  $\mathbb{T}_E$ for manual evaluation, and the results are shown in Table \ref{table4}.

\subsection{Knowledge graph construction in the fine domain}

\begin{figure*}[t]
	\centering
	\includegraphics[width=0.8\textwidth]{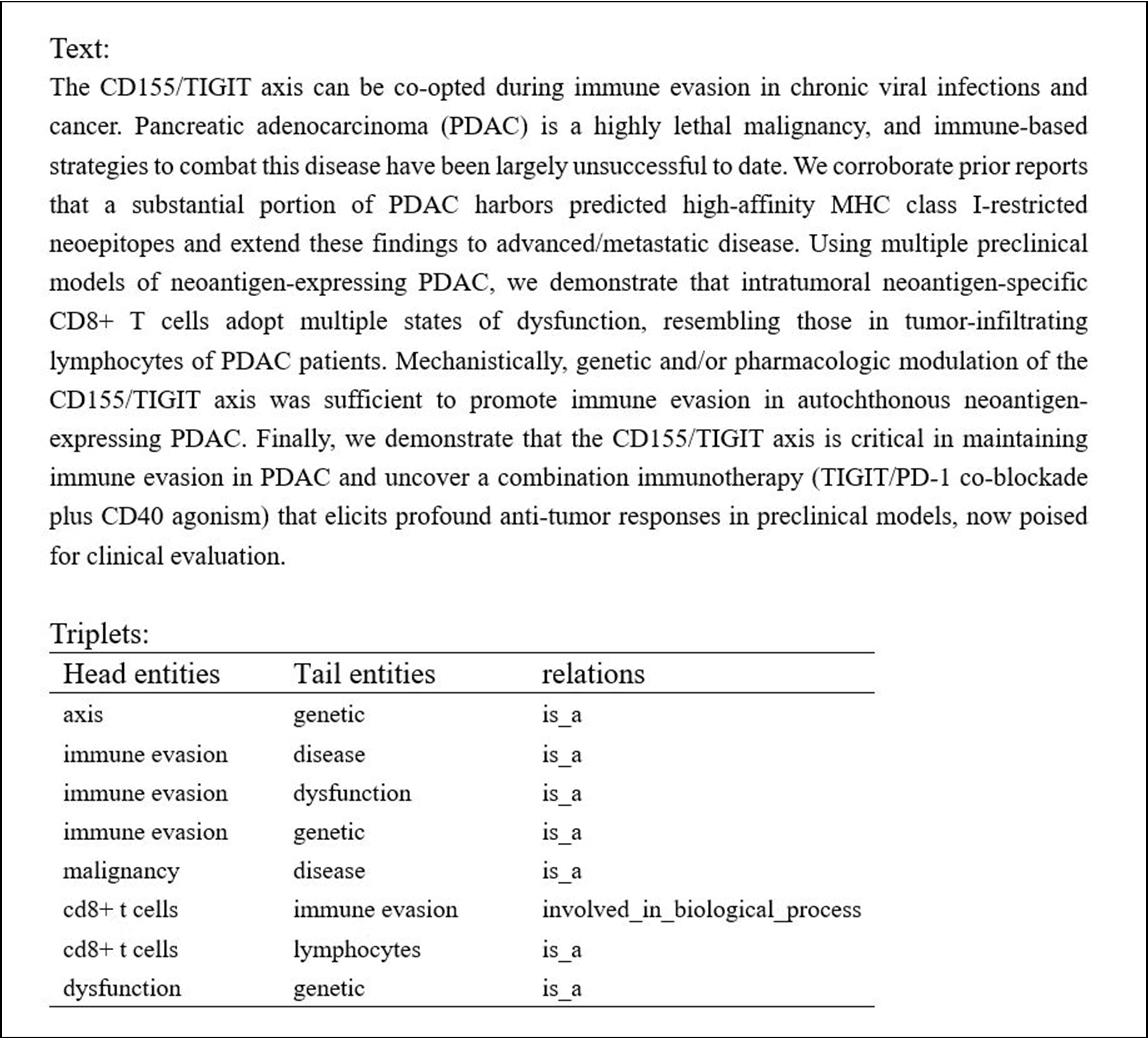} % Reduce the figure size so that it is slightly narrower than the column.
	\caption{Text from a sample abstract used as input to the model testing and the corresponding triplets extracted from the text.}
	\label{fig3}
\end{figure*}

As our ultimate goal, we can construct the KG in the fine domain by combining $\mathbb{T}_{O}$, $\mathbb{T}_R$, and $\mathbb{T}_E$. We selected biomed\_RoBERTa as the backbone language model for KGDA and constructed the knowledge graph correspondingly.  An example of the KG we built is shown in the supplementary material.
Knowledge related to \textit{adenocarcinoma} and \textit{breast cancer} are visualized in Figure \ref{fig2}.

\subsection{Ablation study}

\begin{table}[h]
	\centering
	%\resizebox{.95\columnwidth}{!}{
	\begin{tabular}{lll}
            \hline
            models                     & NER precision      & RE precision      \\ \hline
            \textbf{BERT}              & \textbf{0.908} & \textbf{0.987} \\
            w/o (cumulative)           & 0.888          & 0.985          \\
            w/o (iter)                 & 0.894          & 0.984          \\ \hline
            \textbf{Bio\_ClinicalBERT} & \textbf{0.909} & \textbf{0.988} \\
            w/o (cumulative)           & 0.887          & 0.985          \\
            w/o (iter)                 & 0.898          & 0.985          \\ \hline
            \textbf{biomed\_RoBERTa}   & \textbf{0.908} & \textbf{0.987} \\
            w/o (cumulative)           & 0.888          & 0.983          \\
            w/o (iter)                 & 0.894          & 0.986          \\ \hline
            \end{tabular}
	\caption{Results of ablation study.}
	\label{table5}
\end{table}

We investigated the impact of techniques employed by KGDA on its held-out experiment performance by removing the corresponding component from the framework:

\textbf{w/o (cumulative)}: When using corpus $\mathbb{D}_{i}$ to train NER and RE models, the cumulative corpus is not used. i.e. delete lines 9 and 10 in Algorithm \ref{alg:algorithm1} and mark $corp_N^{'}$ and $corp_R^{'}$ in line 8 as $corp_{N}$ and $corp_{R}$ respectively. 

\textbf{w/o (iter)}: Remove the iterative training strategy and only use $\mathbb{K}_C$ as an external knowledge base.

The results of the ablation analysis are shown in Table \ref{table5}. Compared to the complete framework with w/o (cumulative), it can be seen that the using of accumulated data through iterations is beneficial for improving the generalization ability of NER and RE models. The held-out performances of the model without iteration indicates that the iterative training strategy can not only discover the specific knowledge in the fine domain but also maintain the ability to discover overlapping knowledge between the coarse and fine domain. 

\subsection{Case study}

In this case study, we illustrate a sample case of the model input (text from an abstract) and the model output (triplets extracted from the text) during the testing phase \ref{fig3}. It can be observed from the results that the proposed KGDA framework can recover name entities that are unique in the oncology domain (e.g., immune evasion and cd8+t cells). We have also asked our physicians to review the extracted relations, which are all clinically meaningful and accurate.

\section{Conclusion and Discussion}
In this paper, we propose an integrated, end-to-end framework for knowledge graph domain adaptation using distant supervision, which can be used to construct KG from fully unlabeled raw text data with the guidance of an existing KG. To deal with the potential challenges in distant supervision, which might limit the knowledge discovered from the new domain, we propose an iterative training strategy, which divides an unlabeled corpus into multiple corpuses. For each new corpus to the model, we then combine the knowledge in the coarse domain with the knowledge identified from the previous corpuses for distantly-supervised training. By adopting the iterative training strategy, our proposed KGDA framework can discover not only knowledge that overlaps with the coarse domain, but also knowledge specific to the fine domain and unknown to the coarse domain, thus enabling coarse-to-fine domain adaptation. We implemented the adaptation from biomedical KG to the oncology domain in our experiments and verified the effectiveness of the KGDA framework through held-out and manual evaluation.

Several limitations and challenges remain beyond the current work for more effective and accurate KG construction: Firstly, more thorough evaluation with human reader study is needed to validate that new knowledge relevant (not only correct) to the target domain can be discovered by KGDA. Secondly, it has been recognized by the field that distant supervision will inevitably introduce noisy labels \cite{liang2020bond,zhang2021readsre}, thus the denoising step is usually needed but not implemented in the current version of KGDA. Thirdly, there has been existing KG constructed in the related domains of oncology and cancer research. We will investigate the scheme to allow adaption from multiple sources (not only the coarse domain) to leverage this existing knowledge better. Another type of crucial prior information for this work is clinical ontology, where we will integrate the relationships defined in ontology and entity description to enhance the model. Fourthly, an essential premise of the KGDA is that we assume the source and target domains share the same set of entity types and relation types, which can limit the knowledge discovered from the fine domain. We will investigate data mining techniques to adaptively add/remove entity and relation types in the fine domain. Finally, there have been many new large-scale pre-trained language models developed such as GPT-3 in recent years. While our model uses variations of BERT (BiomedRoBERTa and BioClinicalBERT) as backbone networks, we can easily adapt KGDA to other language models.

% \begin{verbatim}
%   \bibliographystyle{ACM-Reference-Format}
%   \bibliography{conference}
% \end{verbatim}

\bibliography{conference}


\begin{thebibliography}{35}

%%% the Web address.

\ifx \showCODEN    \undefined \def \showCODEN     #1{\unskip}     \fi
\ifx \showDOI      \undefined \def \showDOI       #1{#1}\fi
\ifx \showISBNx    \undefined \def \showISBNx     #1{\unskip}     \fi
\ifx \showISBNxiii \undefined \def \showISBNxiii  #1{\unskip}     \fi
\ifx \showISSN     \undefined \def \showISSN      #1{\unskip}     \fi
\ifx \showLCCN     \undefined \def \showLCCN      #1{\unskip}     \fi
\ifx \shownote     \undefined \def \shownote      #1{#1}          \fi
\ifx \showarticletitle \undefined \def \showarticletitle #1{#1}   \fi
\ifx \showURL      \undefined \def \showURL       {\relax}        \fi
% invisible to TeX
\providecommand\bibfield[2]{#2}
\providecommand\bibinfo[2]{#2}
\providecommand\natexlab[1]{#1}
\providecommand\showeprint[2][]{arXiv:#2}

\bibitem[Alsentzer et~al\mbox{.}(2019)]%
        {alsentzer2019publicly}
\bibfield{author}{\bibinfo{person}{Emily Alsentzer}, \bibinfo{person}{John~R
  Murphy}, \bibinfo{person}{Willie Boag}, \bibinfo{person}{Wei-Hung Weng},
  \bibinfo{person}{Di Jin}, \bibinfo{person}{Tristan Naumann},
  \bibinfo{person}{WA Redmond}, {and} \bibinfo{person}{Matthew~BA McDermott}.}
  \bibinfo{year}{2019}\natexlab{}.
\newblock \showarticletitle{Publicly Available Clinical BERT Embeddings}.
\newblock \bibinfo{journal}{\emph{NAACL HLT 2019}} (\bibinfo{year}{2019}),
  \bibinfo{pages}{72}.
\newblock


\bibitem[Angeli et~al\mbox{.}(2015)]%
        {angeli2015leveraging}
\bibfield{author}{\bibinfo{person}{Gabor Angeli}, \bibinfo{person}{Melvin
  Jose~Johnson Premkumar}, {and} \bibinfo{person}{Christopher~D Manning}.}
  \bibinfo{year}{2015}\natexlab{}.
\newblock \showarticletitle{Leveraging linguistic structure for open domain
  information extraction}. In \bibinfo{booktitle}{\emph{Proceedings of the 53rd
  Annual Meeting of the Association for Computational Linguistics and the 7th
  International Joint Conference on Natural Language Processing (Volume 1: Long
  Papers)}}. \bibinfo{pages}{344--354}.
\newblock


\bibitem[Chen et~al\mbox{.}(2021)]%
        {chen2021explicitly}
\bibfield{author}{\bibinfo{person}{Pei Chen}, \bibinfo{person}{Haibo Ding},
  \bibinfo{person}{Jun Araki}, {and} \bibinfo{person}{Ruihong Huang}.}
  \bibinfo{year}{2021}\natexlab{}.
\newblock \showarticletitle{Explicitly Capturing Relations between Entity
  Mentions via Graph Neural Networks for Domain-specific Named Entity
  Recognition}. In \bibinfo{booktitle}{\emph{Proceedings of the 59th Annual
  Meeting of the Association for Computational Linguistics and the 11th
  International Joint Conference on Natural Language Processing (Volume 2:
  Short Papers)}}.
\newblock


\bibitem[Fader et~al\mbox{.}(2011)]%
        {fader2011identifying}
\bibfield{author}{\bibinfo{person}{Anthony Fader}, \bibinfo{person}{Stephen
  Soderland}, {and} \bibinfo{person}{Oren Etzioni}.}
  \bibinfo{year}{2011}\natexlab{}.
\newblock \showarticletitle{Identifying relations for open information
  extraction}. In \bibinfo{booktitle}{\emph{Proceedings of the 2011 conference
  on empirical methods in natural language processing}}.
  \bibinfo{pages}{1535--1545}.
\newblock


\bibitem[Fries et~al\mbox{.}(2017)]%
        {fries2017swellshark}
\bibfield{author}{\bibinfo{person}{Jason Fries}, \bibinfo{person}{Sen Wu},
  \bibinfo{person}{Alex Ratner}, {and} \bibinfo{person}{Christopher R{\'e}}.}
  \bibinfo{year}{2017}\natexlab{}.
\newblock \showarticletitle{Swellshark: A generative model for biomedical named
  entity recognition without labeled data}.
\newblock \bibinfo{journal}{\emph{arXiv preprint arXiv:1704.06360}}
  (\bibinfo{year}{2017}).
\newblock


\bibitem[Gururangan et~al\mbox{.}(2020)]%
        {gururangan2020don}
\bibfield{author}{\bibinfo{person}{Suchin Gururangan}, \bibinfo{person}{Ana
  Marasovi{\'c}}, \bibinfo{person}{Swabha Swayamdipta}, \bibinfo{person}{Kyle
  Lo}, \bibinfo{person}{Iz Beltagy}, \bibinfo{person}{Doug Downey}, {and}
  \bibinfo{person}{Noah~A Smith}.} \bibinfo{year}{2020}\natexlab{}.
\newblock \showarticletitle{Don’t Stop Pretraining: Adapt Language Models to
  Domains and Tasks}. In \bibinfo{booktitle}{\emph{Proceedings of the 58th
  Annual Meeting of the Association for Computational Linguistics}}.
  \bibinfo{pages}{8342--8360}.
\newblock


\bibitem[Hochreiter and Schmidhuber(1997)]%
        {hochreiter1997long}
\bibfield{author}{\bibinfo{person}{Sepp Hochreiter} {and}
  \bibinfo{person}{J{\"u}rgen Schmidhuber}.} \bibinfo{year}{1997}\natexlab{}.
\newblock \showarticletitle{Long short-term memory}.
\newblock \bibinfo{journal}{\emph{Neural computation}} \bibinfo{volume}{9},
  \bibinfo{number}{8} (\bibinfo{year}{1997}), \bibinfo{pages}{1735--1780}.
\newblock


\bibitem[Jia et~al\mbox{.}(2020)]%
        {jia2020entity}
\bibfield{author}{\bibinfo{person}{Chen Jia}, \bibinfo{person}{Yuefeng Shi},
  \bibinfo{person}{Qinrong Yang}, {and} \bibinfo{person}{Yue Zhang}.}
  \bibinfo{year}{2020}\natexlab{}.
\newblock \showarticletitle{Entity enhanced BERT pre-training for Chinese NER}.
  In \bibinfo{booktitle}{\emph{Proceedings of the 2020 Conference on Empirical
  Methods in Natural Language Processing (EMNLP)}}.
  \bibinfo{pages}{6384--6396}.
\newblock


\bibitem[Kenton and Toutanova(2019)]%
        {kenton2019bert}
\bibfield{author}{\bibinfo{person}{Jacob Devlin Ming-Wei~Chang Kenton} {and}
  \bibinfo{person}{Lee~Kristina Toutanova}.} \bibinfo{year}{2019}\natexlab{}.
\newblock \showarticletitle{BERT: Pre-training of Deep Bidirectional
  Transformers for Language Understanding}. In
  \bibinfo{booktitle}{\emph{Proceedings of NAACL-HLT}}.
  \bibinfo{pages}{4171--4186}.
\newblock


\bibitem[Kertkeidkachorn and Ichise(2017)]%
        {kertkeidkachorn2017t2kg}
\bibfield{author}{\bibinfo{person}{Natthawut Kertkeidkachorn} {and}
  \bibinfo{person}{Ryutaro Ichise}.} \bibinfo{year}{2017}\natexlab{}.
\newblock \showarticletitle{T2kg: An end-to-end system for creating knowledge
  graph from unstructured text}. In \bibinfo{booktitle}{\emph{Workshops at the
  Thirty-First AAAI Conference on Artificial Intelligence}}.
\newblock


\bibitem[Li et~al\mbox{.}(2022b)]%
        {li2022lp}
\bibfield{author}{\bibinfo{person}{Da Li}, \bibinfo{person}{Ming Yi}, {and}
  \bibinfo{person}{Yukai He}.} \bibinfo{year}{2022}\natexlab{b}.
\newblock \showarticletitle{LP-BERT: Multi-task Pre-training Knowledge Graph
  BERT for Link Prediction}.
\newblock \bibinfo{journal}{\emph{arXiv preprint arXiv:2201.04843}}
  (\bibinfo{year}{2022}).
\newblock


\bibitem[Li et~al\mbox{.}(2022a)]%
        {li2022medukg}
\bibfield{author}{\bibinfo{person}{Nan Li}, \bibinfo{person}{Qiang Shen},
  \bibinfo{person}{Rui Song}, \bibinfo{person}{Yang Chi}, {and}
  \bibinfo{person}{Hao Xu}.} \bibinfo{year}{2022}\natexlab{a}.
\newblock \showarticletitle{MEduKG: A Deep-Learning-Based Approach for
  Multi-Modal Educational Knowledge Graph Construction}.
\newblock \bibinfo{journal}{\emph{Information}} \bibinfo{volume}{13},
  \bibinfo{number}{2} (\bibinfo{year}{2022}), \bibinfo{pages}{91}.
\newblock


\bibitem[Li et~al\mbox{.}(2012)]%
        {li2012joint}
\bibfield{author}{\bibinfo{person}{Qi Li}, \bibinfo{person}{Haibo Li},
  \bibinfo{person}{Heng Ji}, \bibinfo{person}{Wen Wang}, \bibinfo{person}{Jing
  Zheng}, {and} \bibinfo{person}{Fei Huang}.} \bibinfo{year}{2012}\natexlab{}.
\newblock \showarticletitle{Joint bilingual name tagging for parallel corpora}.
  In \bibinfo{booktitle}{\emph{Proceedings of the 21st ACM international
  conference on Information and knowledge management}}.
  \bibinfo{pages}{1727--1731}.
\newblock


\bibitem[Liang et~al\mbox{.}(2020)]%
        {liang2020bond}
\bibfield{author}{\bibinfo{person}{Chen Liang}, \bibinfo{person}{Yue Yu},
  \bibinfo{person}{Haoming Jiang}, \bibinfo{person}{Siawpeng Er},
  \bibinfo{person}{Ruijia Wang}, \bibinfo{person}{Tuo Zhao}, {and}
  \bibinfo{person}{Chao Zhang}.} \bibinfo{year}{2020}\natexlab{}.
\newblock \showarticletitle{Bond: Bert-assisted open-domain named entity
  recognition with distant supervision}. In
  \bibinfo{booktitle}{\emph{Proceedings of the 26th ACM SIGKDD International
  Conference on Knowledge Discovery \& Data Mining}}.
  \bibinfo{pages}{1054--1064}.
\newblock


\bibitem[Lindberg et~al\mbox{.}(1993)]%
        {lindberg1993unified}
\bibfield{author}{\bibinfo{person}{Donald~AB Lindberg},
  \bibinfo{person}{Betsy~L Humphreys}, {and} \bibinfo{person}{Alexa~T McCray}.}
  \bibinfo{year}{1993}\natexlab{}.
\newblock \showarticletitle{The unified medical language system}.
\newblock \bibinfo{journal}{\emph{Yearbook of medical informatics}}
  \bibinfo{volume}{2}, \bibinfo{number}{01} (\bibinfo{year}{1993}),
  \bibinfo{pages}{41--51}.
\newblock


\bibitem[Manning et~al\mbox{.}(2014)]%
        {manning2014stanford}
\bibfield{author}{\bibinfo{person}{Christopher~D Manning},
  \bibinfo{person}{Mihai Surdeanu}, \bibinfo{person}{John Bauer},
  \bibinfo{person}{Jenny~Rose Finkel}, \bibinfo{person}{Steven Bethard}, {and}
  \bibinfo{person}{David McClosky}.} \bibinfo{year}{2014}\natexlab{}.
\newblock \showarticletitle{The Stanford CoreNLP natural language processing
  toolkit}. In \bibinfo{booktitle}{\emph{Proceedings of 52nd annual meeting of
  the association for computational linguistics: system demonstrations}}.
  \bibinfo{pages}{55--60}.
\newblock


\bibitem[Mehta et~al\mbox{.}(2019)]%
        {mehta2019scalable}
\bibfield{author}{\bibinfo{person}{Aman Mehta}, \bibinfo{person}{Aashay
  Singhal}, {and} \bibinfo{person}{Kamalakar Karlapalem}.}
  \bibinfo{year}{2019}\natexlab{}.
\newblock \showarticletitle{Scalable knowledge graph construction over text
  using deep learning based predicate mapping}. In
  \bibinfo{booktitle}{\emph{Companion Proceedings of The 2019 World Wide Web
  Conference}}. \bibinfo{pages}{705--713}.
\newblock


\bibitem[Mintz et~al\mbox{.}(2009)]%
        {mintz2009distant}
\bibfield{author}{\bibinfo{person}{Mike Mintz}, \bibinfo{person}{Steven Bills},
  \bibinfo{person}{Rion Snow}, {and} \bibinfo{person}{Dan Jurafsky}.}
  \bibinfo{year}{2009}\natexlab{}.
\newblock \showarticletitle{Distant supervision for relation extraction without
  labeled data}. In \bibinfo{booktitle}{\emph{Proceedings of the Joint
  Conference of the 47th Annual Meeting of the ACL and the 4th International
  Joint Conference on Natural Language Processing of the AFNLP}}.
  \bibinfo{pages}{1003--1011}.
\newblock


\bibitem[Rezayi et~al\mbox{.}(2022)]%
        {rezayi2022clinicalradiobert}
\bibfield{author}{\bibinfo{person}{Saed Rezayi}, \bibinfo{person}{Haixing Dai},
  \bibinfo{person}{Zhengliang Liu}, \bibinfo{person}{Zihao Wu},
  \bibinfo{person}{Akarsh Hebbar}, \bibinfo{person}{Andrew~H Burns},
  \bibinfo{person}{Lin Zhao}, \bibinfo{person}{Dajiang Zhu},
  \bibinfo{person}{Quanzheng Li}, \bibinfo{person}{Wei Liu}, {et~al\mbox{.}}}
  \bibinfo{year}{2022}\natexlab{}.
\newblock \showarticletitle{ClinicalRadioBERT: Knowledge-Infused Few Shot
  Learning for Clinical Notes Named Entity Recognition}. In
  \bibinfo{booktitle}{\emph{Machine Learning in Medical Imaging: 13th
  International Workshop, MLMI 2022, Held in Conjunction with MICCAI 2022,
  Singapore, September 18, 2022, Proceedings}}. Springer,
  \bibinfo{pages}{269--278}.
\newblock


\bibitem[Rossanez et~al\mbox{.}(2020)]%
        {rossanez2020kgen}
\bibfield{author}{\bibinfo{person}{Anderson Rossanez},
  \bibinfo{person}{Julio~Cesar Dos~Reis}, \bibinfo{person}{Ricardo da~Silva
  Torres}, {and} \bibinfo{person}{H{\'e}l{\`e}ne de Ribaupierre}.}
  \bibinfo{year}{2020}\natexlab{}.
\newblock \showarticletitle{KGen: a knowledge graph generator from biomedical
  scientific literature}.
\newblock \bibinfo{journal}{\emph{BMC medical informatics and decision making}}
  \bibinfo{volume}{20}, \bibinfo{number}{4} (\bibinfo{year}{2020}),
  \bibinfo{pages}{1--24}.
\newblock


\bibitem[Rotmensch et~al\mbox{.}(2017)]%
        {rotmensch2017learning}
\bibfield{author}{\bibinfo{person}{Maya Rotmensch}, \bibinfo{person}{Yoni
  Halpern}, \bibinfo{person}{Abdulhakim Tlimat}, \bibinfo{person}{Steven
  Horng}, {and} \bibinfo{person}{David Sontag}.}
  \bibinfo{year}{2017}\natexlab{}.
\newblock \showarticletitle{Learning a health knowledge graph from electronic
  medical records}.
\newblock \bibinfo{journal}{\emph{Scientific reports}} \bibinfo{volume}{7},
  \bibinfo{number}{1} (\bibinfo{year}{2017}), \bibinfo{pages}{1--11}.
\newblock


\bibitem[Roy and Pan(2021)]%
        {roy2021incorporating}
\bibfield{author}{\bibinfo{person}{Arpita Roy} {and} \bibinfo{person}{Shimei
  Pan}.} \bibinfo{year}{2021}\natexlab{}.
\newblock \showarticletitle{Incorporating medical knowledge in BERT for
  clinical relation extraction}. In \bibinfo{booktitle}{\emph{Proceedings of
  the 2021 Conference on Empirical Methods in Natural Language Processing}}.
  \bibinfo{pages}{5357--5366}.
\newblock


\bibitem[Schmitz et~al\mbox{.}(2012)]%
        {schmitz2012open}
\bibfield{author}{\bibinfo{person}{Michael Schmitz}, \bibinfo{person}{Stephen
  Soderland}, \bibinfo{person}{Robert Bart}, \bibinfo{person}{Oren Etzioni},
  {et~al\mbox{.}}} \bibinfo{year}{2012}\natexlab{}.
\newblock \showarticletitle{Open language learning for information extraction}.
  In \bibinfo{booktitle}{\emph{Proceedings of the 2012 joint conference on
  empirical methods in natural language processing and computational natural
  language learning}}. \bibinfo{pages}{523--534}.
\newblock


\bibitem[Smirnova and Cudr{\'e}-Mauroux(2018)]%
        {smirnova2018relation}
\bibfield{author}{\bibinfo{person}{Alisa Smirnova} {and}
  \bibinfo{person}{Philippe Cudr{\'e}-Mauroux}.}
  \bibinfo{year}{2018}\natexlab{}.
\newblock \showarticletitle{Relation extraction using distant supervision: A
  survey}.
\newblock \bibinfo{journal}{\emph{ACM Computing Surveys (CSUR)}}
  \bibinfo{volume}{51}, \bibinfo{number}{5} (\bibinfo{year}{2018}),
  \bibinfo{pages}{1--35}.
\newblock


\bibitem[Stewart and Liu(2020)]%
        {stewart2020seq2kg}
\bibfield{author}{\bibinfo{person}{Michael Stewart} {and} \bibinfo{person}{Wei
  Liu}.} \bibinfo{year}{2020}\natexlab{}.
\newblock \showarticletitle{Seq2kg: an end-to-end neural model for domain
  agnostic knowledge graph (not text graph) construction from text}. In
  \bibinfo{booktitle}{\emph{Proceedings of the International Conference on
  Principles of Knowledge Representation and Reasoning}},
  Vol.~\bibinfo{volume}{17}. \bibinfo{pages}{748--757}.
\newblock


\bibitem[Thanaki(2017)]%
        {thanaki2017python}
\bibfield{author}{\bibinfo{person}{Jalaj Thanaki}.}
  \bibinfo{year}{2017}\natexlab{}.
\newblock \bibinfo{booktitle}{\emph{Python natural language processing}}.
\newblock \bibinfo{publisher}{Packt Publishing Ltd}.
\newblock


\bibitem[Wei(2021)]%
        {wei2021distantly}
\bibfield{author}{\bibinfo{person}{Siheng Wei}.}
  \bibinfo{year}{2021}\natexlab{}.
\newblock \showarticletitle{Distantly Supervision for Relation Extraction via
  LayerNorm Gated Recurrent Neural Networks}. In \bibinfo{booktitle}{\emph{2021
  2nd International Conference on Computing and Data Science (CDS)}}. IEEE,
  \bibinfo{pages}{94--99}.
\newblock


\bibitem[Yu et~al\mbox{.}(2021)]%
        {yu2021domain}
\bibfield{author}{\bibinfo{person}{Haoze Yu}, \bibinfo{person}{Haisheng Li},
  \bibinfo{person}{Dianhui Mao}, {and} \bibinfo{person}{Qiang Cai}.}
  \bibinfo{year}{2021}\natexlab{}.
\newblock \showarticletitle{A domain knowledge graph construction method based
  on Wikipedia}.
\newblock \bibinfo{journal}{\emph{Journal of Information Science}}
  \bibinfo{volume}{47}, \bibinfo{number}{6} (\bibinfo{year}{2021}),
  \bibinfo{pages}{783--793}.
\newblock


\bibitem[Zahera et~al\mbox{.}(2021)]%
        {zahera2021asset}
\bibfield{author}{\bibinfo{person}{Hamada~M Zahera}, \bibinfo{person}{Stefan
  Heindorf}, {and} \bibinfo{person}{Axel-Cyrille~Ngonga Ngomo}.}
  \bibinfo{year}{2021}\natexlab{}.
\newblock \showarticletitle{ASSET: A Semi-supervised Approach for Entity Typing
  in Knowledge Graphs}. In \bibinfo{booktitle}{\emph{Proceedings of the 11th on
  Knowledge Capture Conference}}. \bibinfo{pages}{261--264}.
\newblock


\bibitem[Zeng et~al\mbox{.}(2017)]%
        {zeng2017lstm}
\bibfield{author}{\bibinfo{person}{Donghuo Zeng}, \bibinfo{person}{Chengjie
  Sun}, \bibinfo{person}{Lei Lin}, {and} \bibinfo{person}{Bingquan Liu}.}
  \bibinfo{year}{2017}\natexlab{}.
\newblock \showarticletitle{LSTM-CRF for drug-named entity recognition}.
\newblock \bibinfo{journal}{\emph{Entropy}} \bibinfo{volume}{19},
  \bibinfo{number}{6} (\bibinfo{year}{2017}), \bibinfo{pages}{283}.
\newblock


\bibitem[Zhang et~al\mbox{.}(2021)]%
        {zhang2021readsre}
\bibfield{author}{\bibinfo{person}{Yue Zhang}, \bibinfo{person}{Hongliang Fei},
  {and} \bibinfo{person}{Ping Li}.} \bibinfo{year}{2021}\natexlab{}.
\newblock \showarticletitle{ReadsRE: Retrieval-Augmented Distantly Supervised
  Relation Extraction}. In \bibinfo{booktitle}{\emph{Proceedings of the 44th
  International ACM SIGIR Conference on Research and Development in Information
  Retrieval}}. \bibinfo{pages}{2257--2262}.
\newblock


\bibitem[Zhang et~al\mbox{.}(2020a)]%
        {ZHANG2020102324}
\bibfield{author}{\bibinfo{person}{Yong Zhang}, \bibinfo{person}{Ming Sheng},
  \bibinfo{person}{Rui Zhou}, \bibinfo{person}{Ye Wang},
  \bibinfo{person}{Guangjie Han}, \bibinfo{person}{Han Zhang},
  \bibinfo{person}{Chunxiao Xing}, {and} \bibinfo{person}{Jing Dong}.}
  \bibinfo{year}{2020}\natexlab{a}.
\newblock \showarticletitle{HKGB: An Inclusive, Extensible, Intelligent,
  Semi-auto-constructed Knowledge Graph Framework for Healthcare with
  Clinicians’ Expertise Incorporated}.
\newblock \bibinfo{journal}{\emph{Information Processing and Management}}
  \bibinfo{volume}{57}, \bibinfo{number}{6} (\bibinfo{year}{2020}),
  \bibinfo{pages}{102324}.
\newblock
\showISSN{0306-4573}
\urldef\tempurl%
\url{https://doi.org/10.1016/j.ipm.2020.102324}
\showDOI{\tempurl}


\bibitem[Zhang et~al\mbox{.}(2020b)]%
        {zhang2020document}
\bibfield{author}{\bibinfo{person}{Zhenyu Zhang}, \bibinfo{person}{Bowen Yu},
  \bibinfo{person}{Xiaobo Shu}, \bibinfo{person}{Tingwen Liu},
  \bibinfo{person}{Hengzhu Tang}, \bibinfo{person}{Wang Yubin}, {and}
  \bibinfo{person}{Li Guo}.} \bibinfo{year}{2020}\natexlab{b}.
\newblock \showarticletitle{Document-level relation extraction with dual-tier
  heterogeneous graph}. In \bibinfo{booktitle}{\emph{Proceedings of the 28th
  International Conference on Computational Linguistics}}.
  \bibinfo{pages}{1630--1641}.
\newblock


\bibitem[Zhao et~al\mbox{.}(2019)]%
        {zhao2019construction}
\bibfield{author}{\bibinfo{person}{Mingxiong Zhao}, \bibinfo{person}{Han Wang},
  \bibinfo{person}{Jin Guo}, \bibinfo{person}{Di Liu}, \bibinfo{person}{Cheng
  Xie}, \bibinfo{person}{Qing Liu}, {and} \bibinfo{person}{Zhibo Cheng}.}
  \bibinfo{year}{2019}\natexlab{}.
\newblock \showarticletitle{Construction of an industrial knowledge graph for
  unstructured chinese text learning}.
\newblock \bibinfo{journal}{\emph{Applied Sciences}} \bibinfo{volume}{9},
  \bibinfo{number}{13} (\bibinfo{year}{2019}), \bibinfo{pages}{2720}.
\newblock


\bibitem[Zheng et~al\mbox{.}(2021)]%
        {zheng2021distantly}
\bibfield{author}{\bibinfo{person}{Honghao Zheng}, \bibinfo{person}{Hongtao
  Yu}, \bibinfo{person}{Yinuo Hao}, \bibinfo{person}{Yiteng Wu}, {and}
  \bibinfo{person}{Shaomei Li}.} \bibinfo{year}{2021}\natexlab{}.
\newblock \showarticletitle{Distantly Supervised Named Entity Recognition with
  Spy-PU Algorithm}. In \bibinfo{booktitle}{\emph{2021 IEEE 2nd International
  Conference on Pattern Recognition and Machine Learning (PRML)}}. IEEE,
  \bibinfo{pages}{56--63}.
\newblock


\end{thebibliography}
\bibliographystyle{ACM-Reference-Format}
\end{document}